\documentclass[conference, a4paper]{ieeeconf}  

\IEEEoverridecommandlockouts                              
\usepackage{graphics} 
\usepackage{epsfig} 
\usepackage{mathptmx} 
\usepackage{times} 
\usepackage{amsmath} 
\usepackage{amssymb}  
\usepackage{caption}
\usepackage{subcaption}
\usepackage{float}

\def\quad{\hskip1em\relax}
\def\qquad{\hskip2em\relax}

\title{\LARGE \bf
Computationally Efficient Approaches for Image Style Transfer
}

\author{Ram Krishna Pandey$^{1}$, Samarjit Karmakar$^{2}$ and A G Ramakrishnan$^{1}$
\thanks{$^{1}$Ram Krishna Pandey and A G Ramakrishnan are with Indian Institute of Science, Bangalore, India
        {ramp@iisc.ac.in; agr@iisc.ac.in}}%
\thanks{$^{2}$Samarjit Karmakar is with National Institute of Technology, Warangal, India.
        {ksamarjit@student.nitw.ac.in}}%
}

\begin{document}

\maketitle
\thispagestyle{empty}
\pagestyle{empty}

\begin{abstract}

In this work, we have investigated various style transfer approaches and (i) examined how the stylized reconstruction changes with the change of loss function and (ii) provided a computationally efficient solution for the same. We have used elegant techniques like depth-wise separable convolution in place of convolution and nearest neighbor interpolation in place of transposed convolution. Further, we have also added multiple interpolations in place of transposed convolution. The results obtained are perceptually similar in quality, while being computationally very efficient. The decrease in the computational complexity of our architecture is validated by the decrease in the testing time by 26.1\%, 39.1\%, and 57.1\%, respectively. 

\end{abstract}

\section{INTRODUCTION}
Art is something which humans perceive possibly different from animals do. From times immemorial, people have been fascinated by art and famous artwork. It reflects, transmits and shapes our culture. History suggests that art is made for the production of ``beauty". Humans are emotionally attached to beauty, perhaps this connection between beauty and emotion is the origin of art. As it is said that beauty lies in the eye of the beholder. Humans have the ability to distinguish between the texture, content, and style of images. With a complex interplay between these features, they can compose fine works of art. But this task is very difficult for a machine to do. Recently, convolutional neural networks have shown promising results in computer vision tasks like object recognition. The ability of convolutional neural networks to captures the information in the hierarchy of representation, from low level (from the raw information of a pixel) to high level (more information about the content of the image) led the developers of artistic neural style transfer to think in this direction, to develop an algorithm for creating artistic images~\cite{6}.

The task of Artistic Style Transfer is interesting in the sense that, the representation of the content and style in the convolutional neural network are separable, therefore we can super-impose the texture or style of an artistic work (image) onto the content of a natural scene image. Artistic style transfer aims at super-imposing the artistic style of an artist's artwork onto an image with the help of a learning algorithm/architecture that requires a good amount of computational power to train such architecture.

\section{Literature Review}

\begin{figure*}[!ht]
\centering
\includegraphics[height=0.30\textheight,width=0.80\textwidth]{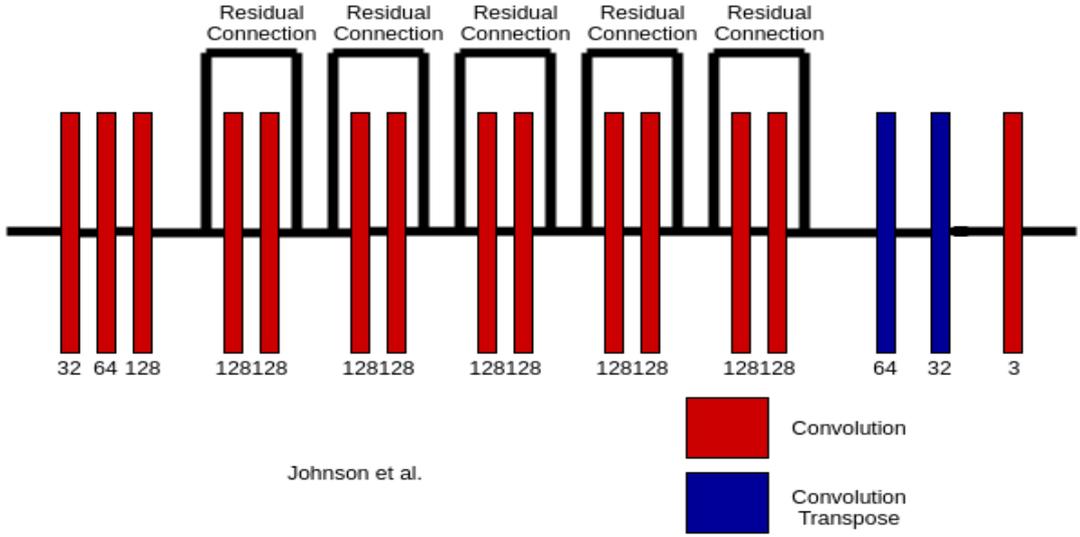}
\caption{The original image transformation network architecture for fast neural transfer as proposed by Johnson et al~\cite{2}. The model consists of convolutional layers, residual blocks and transpose convolutional layers. The first convolutional layer has 32 filters of size 9x9, the second has 64 filters of size 3x3 and the third has 128 filters of size 3x3. The residual blocks contain convolutional layers having 128 filters of size 3x3. The first transpose convolution layer has 64 filters of size 3x3 and the second has 32 filters of size 3x3. The last convolutional layer has 3 filters of size 9x9. `ReLU' non-linearity has been applied to all the layers except the last convolutional layer.}
\label{figure1}
\end{figure*}

\begin{figure*}[!ht]
\centering
\includegraphics[height=0.30\textheight,width=0.80\textwidth]{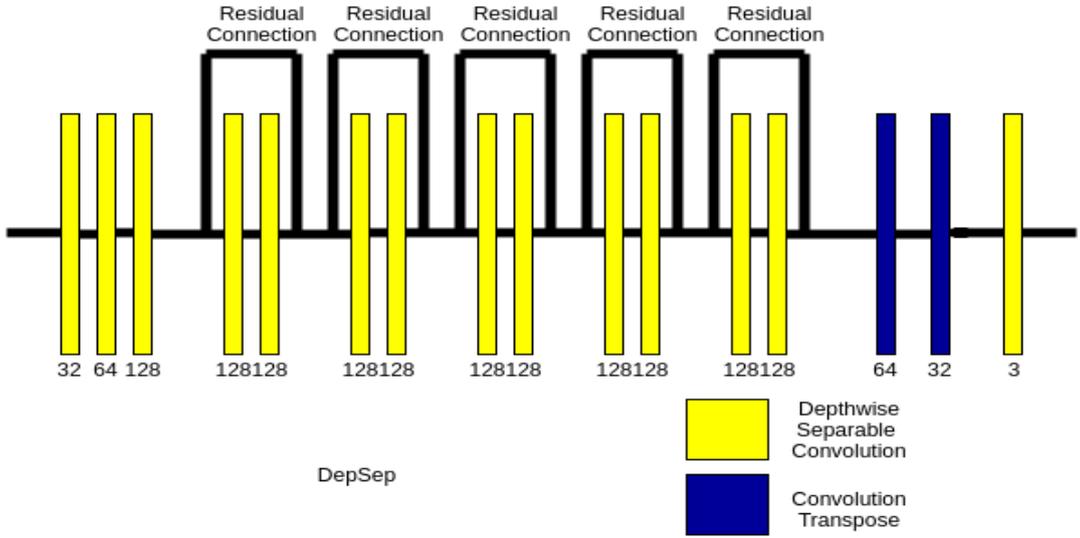}
\caption{The proposed `DepSep' architecture, an improvement over the original image transformation network, by replacing convolutional layers with depth-wise separable convolutional layers. The depthwise separable convolutional layers have channel multiplier = 4.}
\label{figure2}
\end{figure*}

\begin{figure*}
\centering
\includegraphics[height=0.30\textheight,width=0.80\textwidth]{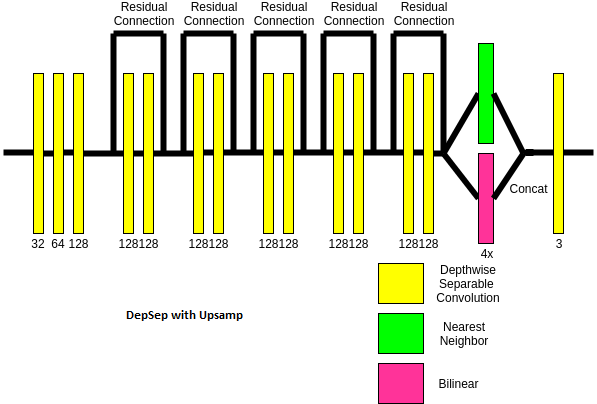}
\caption{The `DepSep with Upsamp' architecture, an improvement over DepSep, by replacing transposed convolutional layers with the concatenation of nearest neighbor and bilinear up-sampling.}
\label{figure3}
\end{figure*}

\begin{figure*}
\centering
\includegraphics[height=0.30\textheight,width=0.80\textwidth]{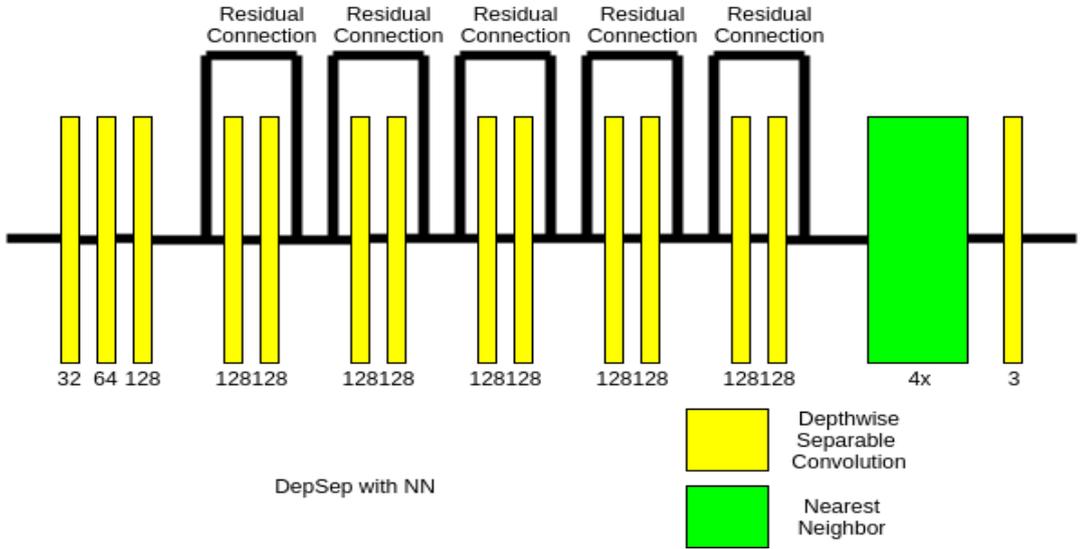}
\caption{The `DepSep with NN' architecture, an improvement over DepSep, by the replacement of transposed convolutional layers with nearest neighbor up-sampling.}
\label{figure4}
\end{figure*}

Gatys et al~\cite{1} was the first to use the power of Convolutional Neural Networks for reproducing famous works of art on natural images. Higher level feature information (content) are captured more in deeper layers. They have shown that the content of an image is the filter responses at the deeper layers of the VGG network~\cite{vgg} trained for image classification task. Since the network has been trained on ImageNet dataset~\cite{10} for image classification task, this network will also capture the content of an image at the deeper layer of the network.

Further, the style of an image is the linear combination of the Gram matrix of the feature maps taken at different layers of the same network. This will capture the correlation of the feature maps at different layers of the network.

Let the feature maps of the $l^{th}$ layer of the network be $F_l(S)$, where $S$ is the style image, then the Gram-based representation of this layer is given as:
\begin{equation}
    G(F_l(S)) = [F_l(S)][F_l(S]^T
    \label{eq1}
\end{equation}
The weighted combination of the Gram-based representation of multiple layers of the network 
is taken for a representation of style of an image. Assuming we take Gram-based representation for layers $l=1, 2, ...n$, the equation for style representation (Sty) for the style image $S$ can be given as follows:
\begin{equation}
    Sty = \sum_{l=1}^n w_l\times G(F_l(S))
\label{eq2}
\end{equation}
The equation for style loss can be given as follows:
\begin{equation}
    l_{style} = \sum_{l=1}^n w_l\times||G(F_l(Y))-G(F_l(S))||
\label{eq3}
\end{equation}
Here, $w_l$ is the weight assigned to the $l^{th}$ layer, $Y$ is the desired image and $S$ is the style image.

The content representation is simply the feature map of the $k^{th}$ layer of the network. The deeper layers of the network are taken as they capture higher-level features of the image more, which is the representation for content, compared to the shallow layers.

The weighted combination of the content representation of multiple layers of the network
is taken for a representation of the content of an image. Assuming we take content representation for layers $k=1, 2, ...n$, the equation for content representation (Con) for the content image $N$ can be given as follows:
\begin{equation}
    Con = \sum_{k=1}^n w_k\times F_k(N)
\label{eq4}
\end{equation}
The equation for content loss can be given as follows:
\begin{equation}
    l_{content} = \sum_{k=1}^n w_k\times||F_k(Y)-F_k(N)||
\label{eq5}
\end{equation}
Here, $w_k$ is the weight assigned to the $k^{th}$ layer, $Y$ is the desired image and $N$ is the content image.

Generally, we consider only higher layers in the network for content. Here, the author has taken the fourth layer (k = 4) for content representation and, first, second, third, fourth and fifth (l = 1, 2, 3, 4, 5) for style representation, with $w_{k}=1$ and $w_{l}=1$. The weights in all other layers were 0.

The objective is to find $\hat{Y}$ such that it satisfies the following equation:
\begin{equation}
    \hat{Y} = \underset{Y}{\mathrm{argmin}}(\alpha l_{content} + \beta l_{style})
\label{eq6}
\end{equation}
This is a parametric slow method to obtain the desired image $Y$, as for each $\{N, S\}$ pair, we have to iteratively optimize the objective function.

Johnson et al~\cite{2} proposed a method for fast neural style transfer, where they train an image transformation network to find a non-linear mapping function which maps from content image to desired output image for a given style image on which the network is trained. They use a loss network pre-trained for image classification task to define perceptual loss functions that measure perceptual differences in content and style between images. The loss network remains fixed during the training process.

The image transformation network is a deep residual convolutional neural network with parameters $\lambda$. It consists of three convolution layers, followed by five residual layers and two transpose convolution layers. The residual networks make it easy in finding the identity function and improving the gradient flow.

The loss network is a pre-trained VGG network on image classification task. They obtain the loss in a similar way obtained by Gatys et al~\cite{1} where they take summation of the feature responses of $k$ layers of the network along with the Gram-based representation of $l$ layers of the network. Along with the content loss and style loss, they also take into account total variation loss, making use of the total variation regularizer which encourages spatial smoothness in the output image. The total loss is given by
\begin{equation}
l_{total} =  \alpha l_{content}+\beta l_{style}+\gamma l_{total variation}
\label{eq7}
\end{equation}
is minimised by backpropagating using Stochastic Gradient Descent optimizer.

This is a computationally faster method for Neural Style Transfer, where they train a feedforward network for a single style image and multiple content images. The output can be obtained, while testing, in a single forward pass as the network, has been trained to stylize any natural image with a single stylized artwork.
\section{Problem Definition and formulation}
\begin{figure*}[!htbp]
\centering
\centering
        \begin{subfigure}[b]{0.33\textwidth}
        \centering
                \includegraphics[height=0.25\textheight,width=0.98\textwidth]{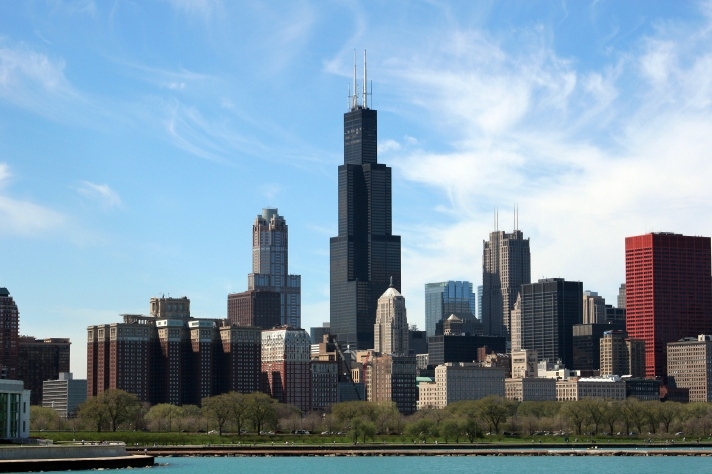}
                \caption{Chicago skyline (content image)}
                \label{fig:gull} 
        \end{subfigure}%
        \begin{subfigure}[b]{0.33\textwidth}
        \centering
                \includegraphics[height=0.25\textheight,width=0.98\textwidth]{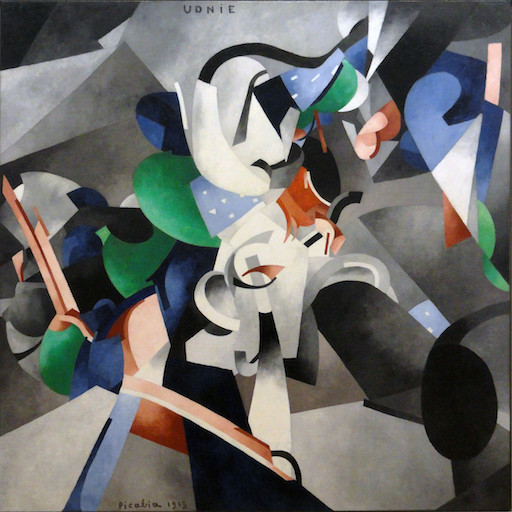}
                \caption{Udnie (style image)}
                \label{fig:gull2}
        \end{subfigure}%
                \begin{subfigure}[b]{0.33\textwidth}
        \centering
                \includegraphics[height=0.25\textheight,width=0.98\textwidth]{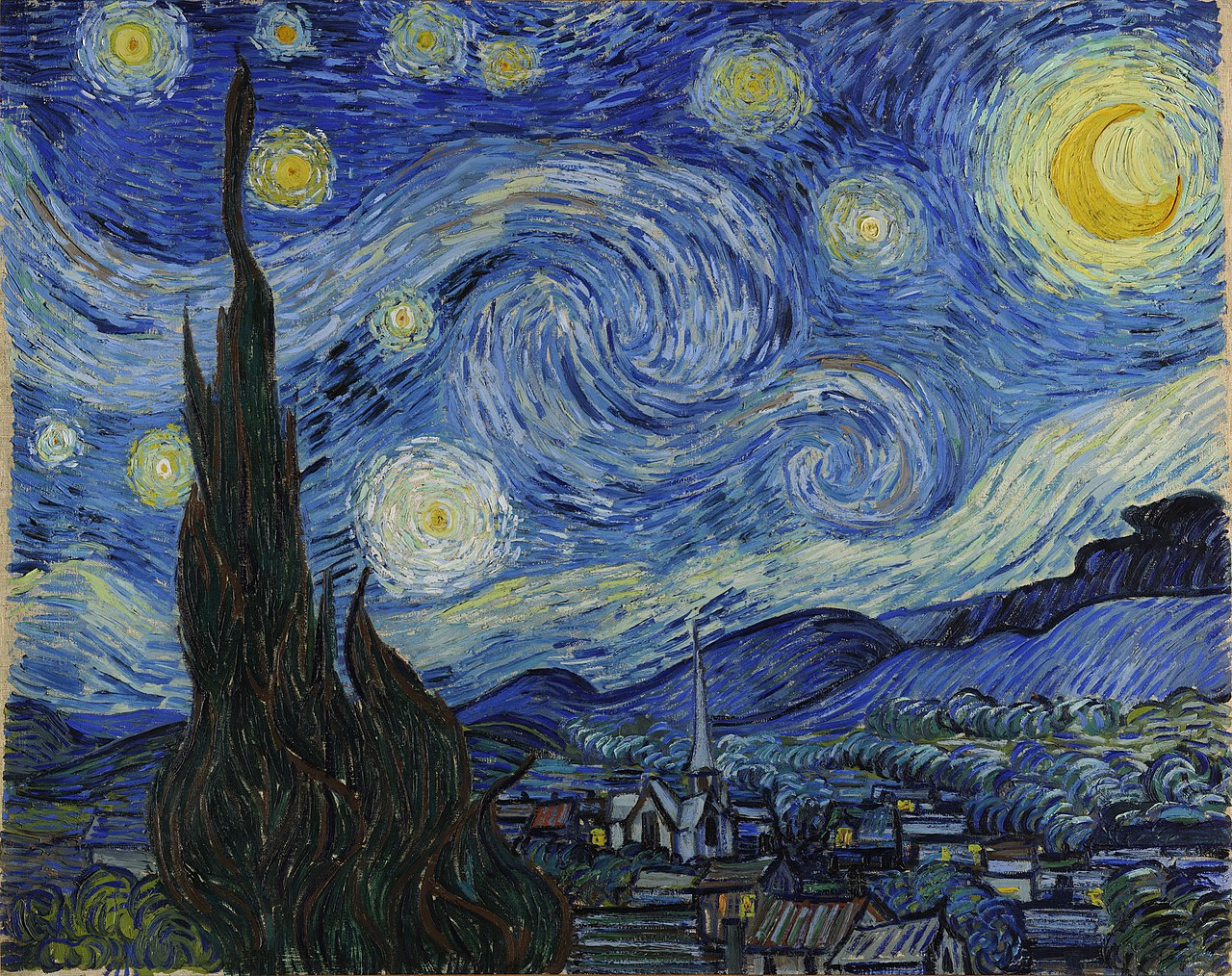}
                \caption{The Starry Night (style image)}
                \label{fig:gull2}
        \end{subfigure}
        \caption{The content image is fed into the model while testing. Each model has been tuned to a specific style image. Hence, for each style image, we must train a model on a dataset of content images. (a) The natural scene image of the Chicago skyline. (b) The "Udnie" image by Francis Picabia, 1913. (c) The "The Starry Night" image by Vincent van Gogh, 1889.}
        \label{figure5}
\end{figure*}

\begin{figure*}[!htbp]
\centering
        \begin{subfigure}[b]{0.25\textwidth}
                \includegraphics[height=0.20\textheight,width=0.95\textwidth]{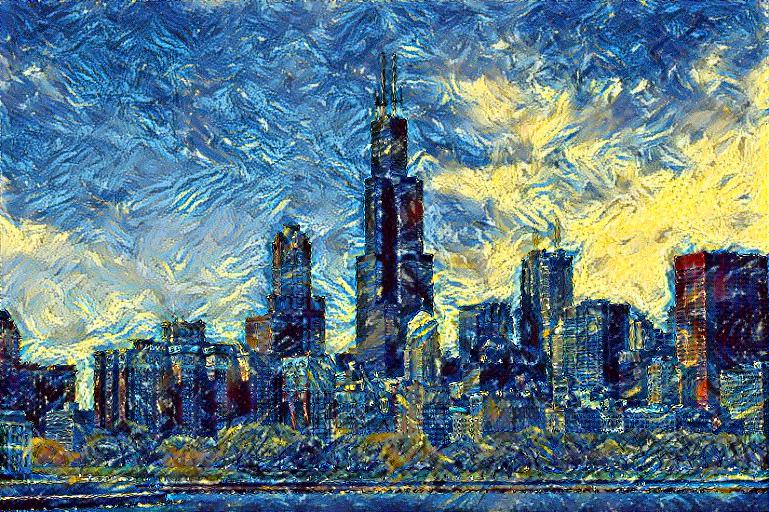}
                \caption{Content loss - MSE\\ \hspace*{0.5cm}Style loss - MSE}
                \label{fig:gull}
        \end{subfigure}%
       \begin{subfigure}[b]{0.25\textwidth}
                \includegraphics[height=0.20\textheight,width=0.95\textwidth]{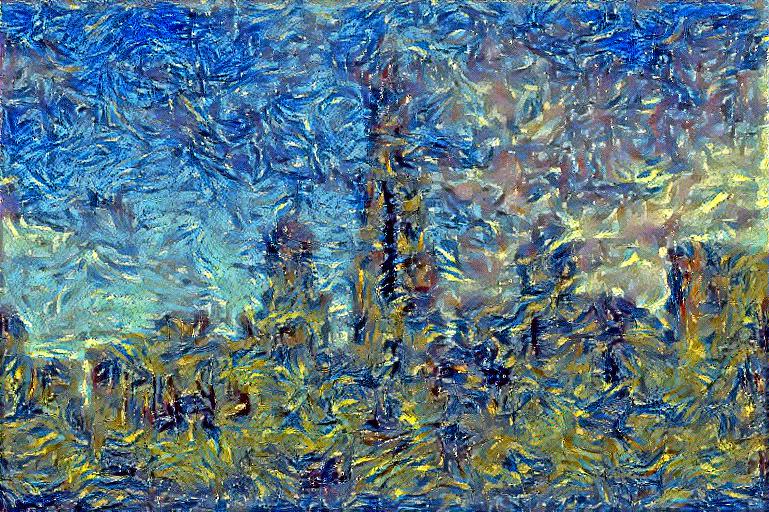}
                \caption{Content loss - Char\\
                \hspace*{0.5cm}Style loss - Char}
                \label{fig:gull2}
        \end{subfigure}%
        \begin{subfigure}[b]{0.25\textwidth}
                \includegraphics[height=0.20\textheight,width=0.95\textwidth]{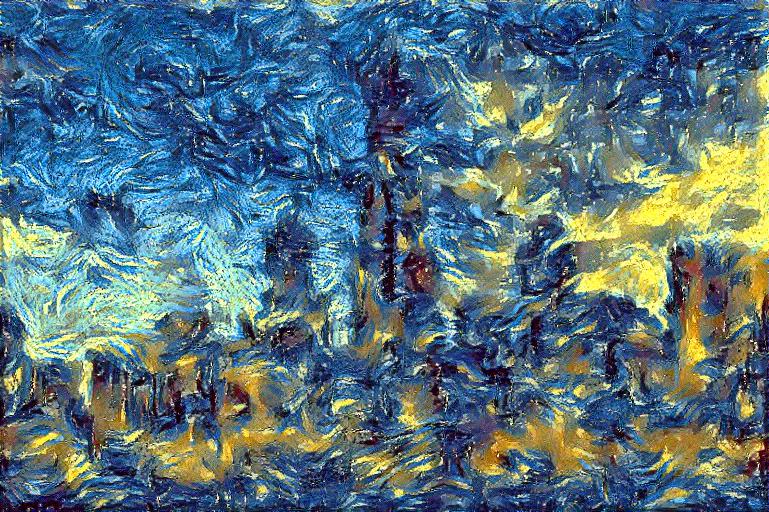}
                \caption{Content loss - MSE\\
                \hspace*{0.5cm}Style loss - Char}
                \label{fig:tiger}
        \end{subfigure}%
        \begin{subfigure}[b]{0.25\textwidth}
                \includegraphics[height=0.20\textheight,width=0.95\textwidth]{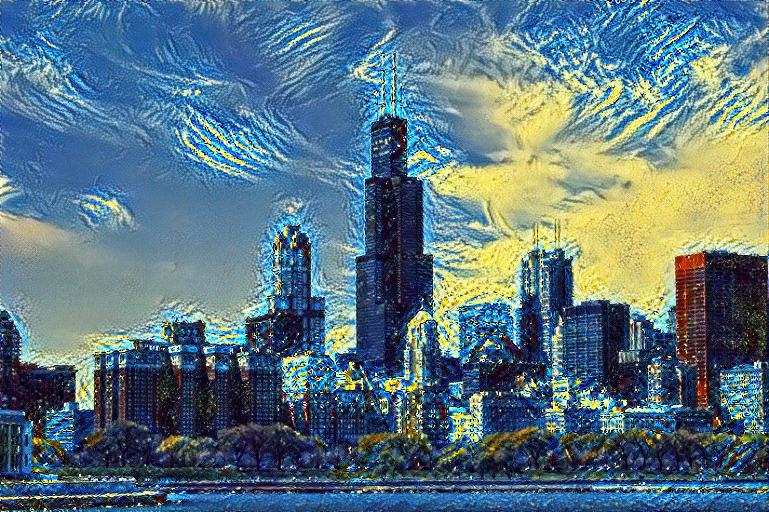}
                \caption{Content loss - Char \\ \hspace*{0.5cm}Style loss - MSE}
                \label{fig:mouse}
        \end{subfigure}
        \caption{A qualitative analysis of the effect of different losses on the stylized output for the method proposed by Gatys et al. We have taken Chicago skyline as the content image and Starry night as the style image. 
        We have varied the content and style losses with MSE loss function and Charbonnier (Char) loss function. Details are mentioned in subcaption above.}
        \label{figure6}
        
\end{figure*}

\begin{figure*}[!h]

	\begin{subfigure}[b]{0.25\textwidth}
                
                \includegraphics[height=0.20\textheight,width=0.95\textwidth]{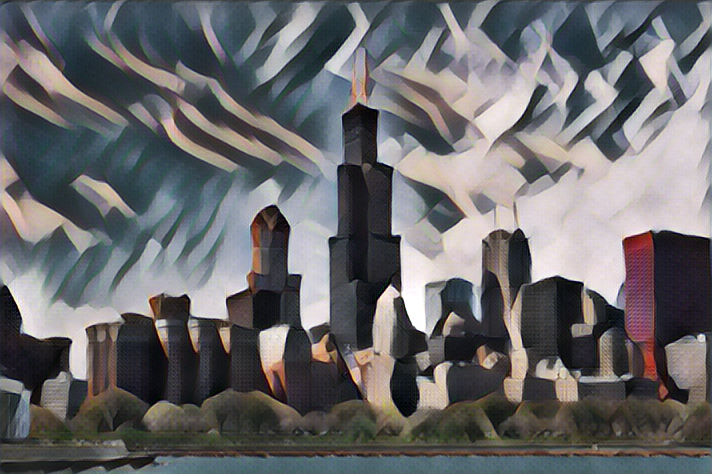}
                \caption{Johnson et al}
                \label{fig:mouse}
        \end{subfigure}%
        \begin{subfigure}[b]{0.25\textwidth}
       \includegraphics[height=0.20\textheight,width=0.95\textwidth]{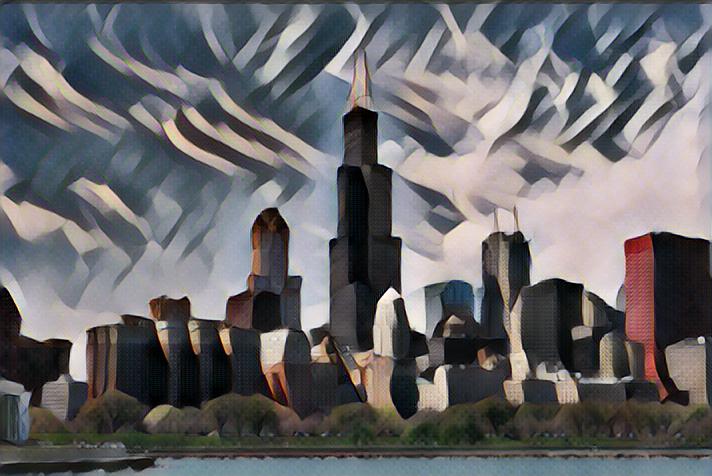}
                \caption{DepSep}
                \label{fig:tiger}
        \end{subfigure}%
         \begin{subfigure}[b]{0.25\textwidth}
                \includegraphics[height=0.20\textheight,width=0.95\textwidth]{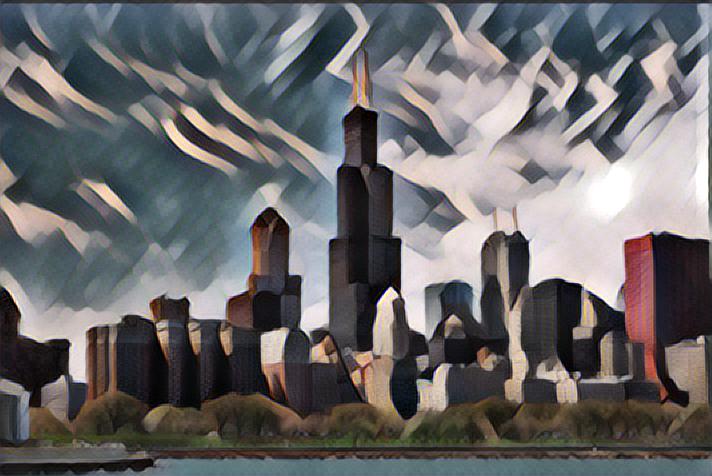}
                \caption{DepSep with Upsamp}
               \label{fig:mouse}
        \end{subfigure}%
       \begin{subfigure}[b]{0.25\textwidth}
                \includegraphics[height=0.20\textheight,width=0.95\textwidth]{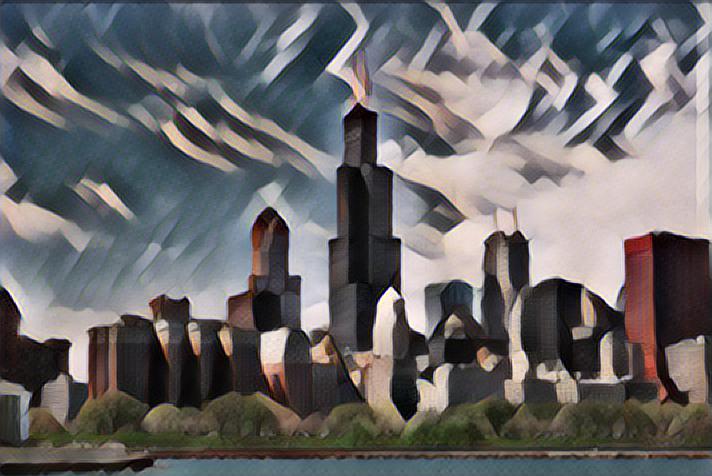}
                \caption{DepSep with NN}
                \label{fig:mouse}
        \end{subfigure}
        \caption{A qualitative analysis of the results obtained from the various architectures proposed. (a) The stylized output from the model proposed by Johnson et al.[2] (b) The stylized output from the model in Fig. 2. (c) The stylized output from the model in Fig. 3. (d) The stylized output from the model in Fig. 4. Chicago skyline is the content image and Udnie is the style image used.
        }
        \label{figure7}
        
\end{figure*}

We aim to solve the above-mentioned problem in a computationally efficient way without much decrease in perceptual quality of the stylized output image.

The problem of neural style transfer can be mathematically formulated as follows:

Suppose $N$ is a given natural scene image (content image), whose content we want to preserve and an artistic work (style image) $S$ whose texture or style we want to super-impose on the content image. Let the desired image obtained be $Y$. We require $\hat{Y}$ which satisfies the following equation:

\begin{equation}
    \hat{Y} = \underset{Y}{\mathrm{argmin}}(\alpha\times||C(Y) - C(N)|| + \beta\times||X(Y) - X(S)||)
    \label{eq8}
\end{equation}

Here, $C$ is an operator which extracts the content of the image and $X$ is an operator which extracts the style of the image. $\hat{Y}$ is the value of $Y$ which minimizes the above objective function.

\section{Contribution}

Our main contributions are as follows:
\begin{itemize}
\item We have tried to preserve more content in the reconstructed image by using Charbonnier loss~\cite{amorerobustloss} to compute the content loss instead of using MSE loss. 
\item On experimentation, Charbonnier loss has been found to preserve more content when used as a replacement for MSE loss for computation of the content loss. Results reported in Figure~\ref{figure6} validate these points.

\item  We have improved the architecture proposed by Johnson et al.~\cite{2} and made it computationally efficient by reducing the number of parameters using depth-wise separable convolution layers instead of regular convolution layers in the image transformation network. Further, we have replaced the deconvolution layer (transpose convolution) with nearest neighbour up-sampling with gradient flow (as shown in Figure~\ref{figure4}) and also replaced deconvolution with the concatenation of nearest neighbour and bilinear up-sampling (as shown in Figure~\ref{figure3}).

\item In Xception model Francois Chollet~\cite{5}  shows the working of depthwise separable convolution in deep learning architectures. The use of depthwise separable convolution results in a huge reduction in the number of parameters but with similar model performance.

\item Nearest Neighbour interpolation is a classical technique and is computationally very efficient method for upsampling images, which repeats neighboring pixel four times to obtain the output image. Results reported in the Table~\ref{table1} validate the above statements of computational efficiency of our models.

\end{itemize}

\section{Our approach}
Our algorithms/architectures for fast style transfer are inspired by the image transformation network proposed by Johnson et al~\cite{2}. All the convolutional layers have been replaced with depth-wise separable convolutional layers. Further, we have replaced fractionally strided or transposed convolution layers with nearest neighbour upsampling as well as with concatenation of nearest neighbour and bilinear upsampling as performed in~\cite{8}. A diagrammatic representation of our architectures has been given in Figures [\ref{figure1}-~\ref{figure4}].

We have used $\alpha=7.5$, $\beta=10^2$ and $\gamma=2 \times 10^2$ for the loss function in equation~\ref{eq7}.

We have implemented the proposed architecture as above and trained the networks on the Microsoft COCO dataset~\cite{6}.

We have used Adam optimizer~\cite{3} to minimise the loss, with learning rate as 0.001, $\beta_{1}$ as 0.999 and $\beta_{2}$ as 0.99.

Instance normalization has been applied after every depthwise separable convolutional layer as suggested by Ulyanov et al~\cite{11}.

The models have been implemented in the Tensorflow~\cite{tensorflow} deep learning framework and trained on Nvidia Titan X GPU, which takes roughly 18-20 hours for 2 epochs (each epoch having 20650 iterations).

\section{Results}

\begin{table}
	\centering
 \caption{Comparison of the testing time taken by the original model~\cite{2} with those of the modified architectures proposed by us as shown in Figures~\ref{figure2},~\ref{figure3} and~\ref{figure4}.}
    
	\resizebox{0.48\textwidth}{!}
	{
		\begin{tabular}{|c |c|c|} \hline
			
			 \bf Details 						& \bf Method            & \bf Testing time in sec.  \\ \hline
			
			Fig.~\ref{figure1}                      		&Johnson et al.             					&  1.33   \\ \hline
			
            Fig.~\ref{figure2}           	 	&DepSep 										&  0.97   \\  \hline
	            Fig.~\ref{figure3}       		& DepSep with Upsamp     						&  0.81   \\ \hline
					
            Fig.~\ref{figure4}            		&DepSep with NN                					&  0.57   \\ \hline

		\end{tabular}
	}
   
	\label{table1}
\end{table}
\subsection{Experimental results}
The testing time for the models has been given in Table~\ref{table1}. We see that replacing convolution layers with depth-wise separable convolution layers has led to 26.06\% decrease in testing time. Further, replacing transpose convolution with nearest neighbour upsampling has led to 57.14\% decrease in testing time and replacing it with the concatenation of nearest neighbour and bilinear upsampling has led to 39.09\% decrease in testing time.

Qualitatively, we see from Figure 7, the images produced by all the models are similar with negligible change in perceptual quality.

Hence, we are able to make the model perform faster with negligible change in perceptual quality in the output image.

\subsection{Discussion}
Figures~\ref{figure1}~\ref{figure2}~\ref{figure3} and~\ref{figure4} shows the architecture diagram of the transformation network, originally proposed in~\cite{2} and various modifications done by us respectively. As an initial experiment we first replace all the convolution layer in the original architecture shown in Figure~\ref{figure1} with depthwise separable convolution (shown in Figure~\ref{figure2}) there is around 26\% decrease in testing time. Further, we have replaced the transposed convolution layer with the combination of nearest neighbour and bilinear interpolation as proposed in~\cite{8}   and can be seen in Figure~\ref{figure3}. This gives a decrease in testing time by 39.09\% over the original model shown in Figure~\ref{figure1} and in our last experiment we have replaced the transposed convolution shown in Figure~\ref{figure2} with nearest neighbour interpolation (can be seen in Figure~\ref{figure4}). This gives a decrease in testing time by 57.14\% over the original model.  All the results obtained by these architectures shown in Figures [\ref{figure1}-\ref{figure4}] shows similar results perceptually, but with our techniques shown in Figures [\ref{figure2}-\ref{figure4}], we have decreased the testing time by a huge margin.

Figures~\ref{figure6},~\ref{figure7} and Table~\ref{table1} shows the qualitative and quantitative results of our experiments respectively. The details of which are mentioned in the captions of the Figures and Table. 
\section{CONCLUSIONS}
In the first part of this paper, we have explored various models by changing the loss function in the original implementation proposed by Gatys et al~\cite{1} and have shown their results in Figure~\ref{figure4}. Since there is no metric to decide which of the loss or its combination is better, we have left it for the viewer to decide. This will help researchers working in the similar area to use the proper loss function (or its combination) based on their requirement. Secondly, we have modified the architecture proposed by Johnson et al~\cite{2} in the best possible way using depth-wise separable convolution, nearest neighbour interpolation and the combination of nearest neighbour and bilinear interpolation and have provided three computationally efficient architecture. These proposed architectures can reconstruct the stylized images almost similar in perceptual quality than that reconstructed by the original proposed model in~\cite{2}. Our proposed architecture shows a significant  improvement in testing time by a good margin of 26.06\%, 39.09\%, and 57.14\% respectively.

\end{document}